\pdfoutput=1
\documentclass[11pt]{article}
\usepackage[final]{acl}
\usepackage{times}
\usepackage{latexsym}
\usepackage[T1]{fontenc}
\usepackage[utf8]{inputenc}
\usepackage{microtype}
\usepackage{inconsolata}
\usepackage{graphicx}
\usepackage{subcaption}
\usepackage{amsmath}
\usepackage{booktabs}
\usepackage{multirow}
\usepackage{enumitem}
\usepackage{makecell}
\usepackage{tcolorbox}

\usepackage[font=small,skip=0pt]{caption}      
\usepackage[subrefformat=parens]{subcaption}    
\usepackage[font=small,skip=0pt]{caption}
\definecolor{mypurple}{RGB}{118, 33, 117}
\definecolor{mypink}{RGB}{255, 102, 187}

\title{VIBE: Can a VLM Read the Room?}

\author{Tania Chakraborty, Eylon Caplan, Dan Goldwasser \\
        Purdue University, West Lafayette, IN, USA\\
  \texttt{\{tchakrab,ecaplan,dgoldwas\}@purdue.edu}\\}

\begin{document}
\maketitle
\begin{abstract}
Understanding human social behavior such as recognizing emotions and the social dynamics causing them is an important and challenging problem. While LLMs have made remarkable advances, they are limited to the textual domain and cannot account for the major role that non-verbal cues play in understanding social situations.
Vision Language Models (VLMs) can potentially account for this gap, however their ability to make correct inferences over such social cues has received little attention. In this paper, we explore the capabilities of VLMs at social reasoning. We identify a previously overlooked limitation in VLMs: the Visual Social-Pragmatic Inference gap. To target this gap, we propose a new task for VLMs: Visual Social-Pragmatic Inference. We construct a high quality dataset to test the abilities of a VLM for this task and benchmark the performance of several VLMs on it.
\end{abstract}

\section{Introduction}

\begin{center}
\textit{“It is only with the heart that one can see rightly; what is essential is invisible to the eye.”}
\end{center}
\begin{flushright}
\textit{— \citeauthor{little-prince}, \emph{The Little Prince}}
\end{flushright}

Understanding social dynamics—such as recognizing emotions and their cause—is something humans do intuitively \cite{intuitive-emotions}, yet \emph{how} we do it remains a challenging question. Even emotion perception, an intuitive ability for most people, poses several levels of complexity \cite{emotions-are-complex} arising from a variety of factors such as context, cultural variability and channels of perception (language, vision, etc.). Influential psychology and cognitive neuroscience studies \cite{facs,gelder} have shown that significant proportions of socially relevant information is contained in non-verbal cues like facial expression and body language. In certain situations \cite{mehrabian1967decoding, mehrabian1972nonverbal}, only 7\% of meaning is conveyed through words, with 38\% coming from tone of voice and \emph{55\% through facial expressions!}  Even within the visual modality alone, emotion recognition is not simply a matter of identifying facial expressions \cite{barrett2010context}, as the same facial expression can convey different emotions depending on body posture \cite{Aviezer2012BodyCues}. E.g., even an iconic emotion indicator like a smile, may not always indicate joy \cite{smile-is-not-a-smile}. Despite the intuitive ease with which most people perceive emotions, research has shown that this ability relies on complex mechanisms spanning cognition, perception and understanding.

\begin{figure}[t]
  \centering
  \includegraphics[width=\linewidth]{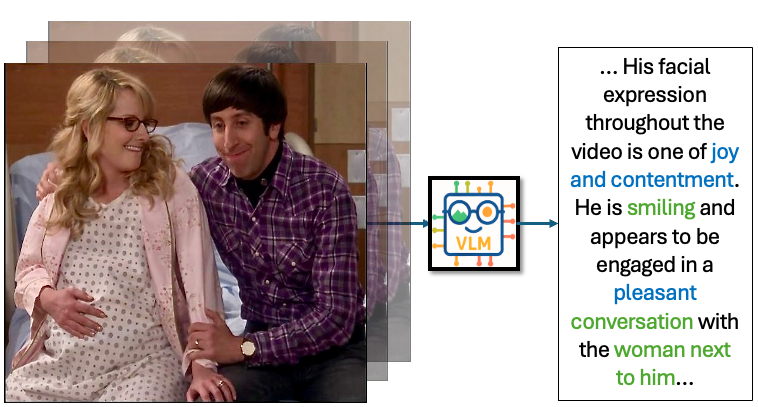}
  \caption{In the video, the man smiles sadly, teary eyed. His partner looks at him with sympathy and pats his leg comfortingly. The VLM (InternVL2 26B) correctly identifies the smile, and the woman next to the man, but is not able to interpret the smile correctly.}
  \label{fig:sad-howard}
\end{figure}

In recent years there has been increasing interest in AI systems that can achieve human levels of social intelligence. LLMs in particular have shown remarkable promise in such tasks, though they struggle when presented with complex scenarios \cite{llm1, llm2, llm3,wu2025identifying} and are limited to the textual domain. The recent emergence of VLMs~\cite{singh2022flava,bai2023qwen,bordes2024introduction,zhang2024vision} provides an exciting opportunity to study social scenes by connecting textual and visual information and leveraging the strong reasoning capabilities of LLMs. However, while some works have explored the efficacy of using VLMs for social intelligence tasks \cite{yang2024emollm,sun2024face,Bhattacharyya2025EvaluatingVM,liu2025culturevlm}, this area still remains relatively underexplored.

In this paper, we conduct a thorough investigation of VLMs' ability to perform social reasoning tasks. We focus on emotion-related inferences drawn from videos that depict a scene's social dynamics. Our goal is to test whether VLMs can \textit{``read the room''}, i.e., assess the alignment between their inferences and human-level understanding of the scene. Our assessment is done at two levels of analysis. First, in terms of the \textit{visual cues} the model identifies as relevant for understanding the social situation, and second in terms of the \textit{pragmatic inferences} made on top of these cues. 
Most works have focused on the correctness of the first step, i.e., testing whether models incorrectly hallucinate a visual pattern~\cite{pope,woodpecker,liu2024survey,favero2024multi,rawte2025defining}. In contrast, we focus on the second and as a result expose a previously overlooked limitation in VLMs: even when the VLM correctly identifies a relevant visual cue, it struggles to interpret it correctly. We refer to this as the \textbf{Visual Social-Pragmatic (VSP) Inference gap}. To further elucidate what we mean by this, we use Figure~\ref{fig:sad-howard} (a frame representing a short video clip) as an example: while a man is clearly smiling, it does \textit{not} indicate joy, as evident by his slumped posture and teary eyes. Furthermore, the second person in the scene (a pregnant woman in hospital gown) looks  at him in sympathy and comforts him. The smile should be interpreted as sad or bittersweet. However, while the VLM correctly identifies the smile, it misinterprets it as evidence for joy.


Our diagnostic process proceeds in three steps. First we conduct a diagnostic analysis over the emotion recognition task in conversational videos (see Sec.~\ref{sec:ep}) by evaluating the contribution of the visual modality. While offering limited (often negative) contribution when fusing the modalities directly, we note an interesting finding - when visual cues derived from the VLM  augment the conversational transcript, they can lead to improved performance. This improvement is not guaranteed and careful tuning is needed, indicating that these cues or inferences made over them are noisy. Second, through human evaluation we separate between noisy cues (i.e., hallucinated) vs. misinterpreted cues (see Sec.~\ref{sec:vsp-analysis}). We curate a dataset, VIBE (VSP Inference of Behavior and Emotion), consisting of 994 unique instances of the VSP Inference task. We benchmark several VLMs on our dataset, and we compare model performances to human performance on the dataset, revealing that humans outperform the best VLM by 17.2\% in accuracy. This indicates that VSP Inference--and not only hallucination--is an important limitation that VLMs struggle with. We perform analyses to elucidate exactly which kinds of social visual cues are hallucinated, misinterpreted, or uninformative. Specifically, we see that the most common failure mode on a downstream emotion recognition task is \emph{misinterpretation}, particularly among subtle facial movements like \emph{gaze and eye behavior} and \emph{furrowed brows} (Sec.~\ref{sec:vsp-analysis}). Our contributions are:

\begin{enumerate}[itemsep=0pt, topsep=0pt]
    \item Expose the VSP inference gap in VLMs. Propose a task and dataset that isolates the gap and provides a tool for measuring it.\footnote{dataset: https://huggingface.co/datasets/taniarini/vibe}
    \item Analyze how the effects of this gap influence performance on a downstream social science task (emotion recognition).
\end{enumerate}

This paper has two main components, presented in Section 4 and Section 5.
\begin{enumerate}
    \item Section 4 (Diagnostic Emotion Prediction Task): This section builds an understanding of current VLM capabilities on tasks that require social common sense and cognitive reasoning and serves as motivation for why this is an important problem. Section 3.2 gives a high level definition and motivation for this task.
    \item Section 5 (Novel VSPI Task): This section describes in detail the construction of the dataset VIBE, and the subsequent results of evaluating several models on it. Section 3.3 provides a high level introduction of the dataset VIBE and the novel VSPI task.
\end{enumerate}



\section{Related Work}
\textbf{Multi-modality in Social Science Tasks:} Many methods have been proposed for multi-modal emotion prediction that involve parametric training  \cite{facial-mmt, telme, yang-etal-2023-self, huang2025emotionqwentraininghybridexperts}. Several other approaches combine LLMs with vision models for multi-modal social understanding, e.g. \citet{SoV, idealGPT, lei2024largevisionlanguagemodelsemotion} propose prompting strategies to work around the limited reasoning of VLMs. Other works use the reasoning powers of an LLM in conjunction with vision tools \cite{smile, visionGPT, Etesam_2024}. Additionally, some datasets have been proposed for exploring models' capabilities in theory of mind and in emotion interpretation \cite{lin2025feelbreakingboundariesemotional, chen2024theorymindseyereading}. The key difference in our work comes from (1) the use of videos for the temporal dimension and (2) the explicit separation that our dataset makes between the problem of hallucination and VSP inference.

\noindent\textbf{Hallucinations in VLMs:} VLMs are known to suffer from hallucinations \cite{liu2024surveyhallucinationlargevisionlanguage}. Many methods attempt to measure and mitigate hallucinations, either by breaking down outputs \cite{pope, woodpecker, petryk2024alohanewmeasurehallucination}, training \cite{ben-kish-etal-2024-mitigating, xie-etal-2024-v}, or decoding algorithms \cite{manevich-tsarfaty-2024-mitigating}. Visual hallucinations fundamentally differ from VSP in that a description or explanation may be factually correct in isolation (identifying a smile), but provide the wrong pragmatic interpretation, and therefore, incorrect meaning (not realizing it is a sad smile).

\noindent\textbf{Pragmatics:} Pragmatics is the study of how context contributes to meaning \cite{theory-of-signs}, and has a rich history in linguistics. Recently, works have sought to understand and improve the pragmatics in LLMs via grounding  of various forms \cite{pub, llm-prag-survey, mohapatra-etal-2024-evaluating, white-etal-2024-communicate}. In the visual domain, VLMs have also been provided with context coming from outside sources, \cite{codis, li-etal-2023-multi-modal, willemsen-etal-2023-resolving}, while other works address generating contrastive captions using pragmatic inferences \cite{contrastive-clip, contrastive-dataset}. However, in the \emph{visual} domain, the study of pragmatics in \emph{social} contexts remains underexplored.

\noindent\textbf{Multimodal Datasets:} There have been excellent datasets that tackle similar problems of social reasoning in multimodal settings. VCR \cite{VCR-dataset} is one such popular image based common-sense dataset. A closely related class of such datasets are image based emotion prediction datasets such as \citet{image-emotion-dataset-1} and \citet{image-emotion-dataset-2}. VIBE differs from these in 2 main respects: (1) in contrast to images, videos add a layer of complexity due to longer context, subtle temporal effects changing the meaning and (2) VIBE tackles a deeper problem than these tasks by specifically testing the abilities of models to \emph{interpret} a visual artifact.

\section{Task Definitions}

In this section we briefly define the two tasks that the paper addresses: The \textit{diagnostic task} of Emotion Prediction (in multi-party conversation videos) and the \textit{novel proposed task} of Visual Social-Pragmatic Inference (VSP Inference). We also define vocabulary used in the paper.

\subsection{Vocabulary}
We define with examples the terms and vocabulary used in this paper.

\textbf{Video Description}: The text description of a video, output by a VLM.

\textbf{Visual Cue:} The text representation of a \textit{directly observable artifact} in a video. E.g. a smile, wave, laughter, etc. Anti-example: happiness (requires inference over what is seen).

\textbf{VSP Inference:} The interpretation of a Visual Cue. Examples:
\begin{enumerate}[itemsep=0pt, topsep=0pt]
    \item \underline{Smile indicating joy:} If everything else about the person aligns with happiness then the VSP inference of joy is correct. (visual cue: smile, VSP inference: joy).
    \item \underline{Smile indicating sadness:} If the person is smiling but also has tears in their eyes and a down-turned mouth, then the VSP inference is that it is a sad smile. (visual cue: smile, VSP inference: sadness).
\end{enumerate}
\textbf{Levels of VSP Inference:} Via prompting, we can `toggle' how much VSP inference a VLM is allowed to insert into its descriptions. We assign a "Level" (1, 3 or 5)\footnote{We do not list (or use in the paper) Levels 2 and 4. They would fall somewhere between their nearest neighbors.} as:
  \begin{enumerate}[itemsep=0pt, topsep=0pt]
      \item Level 1: Only Visual Cues. Eg., The woman had raised eyebrows and pursed lips.
      \item Level 3: Visual Cues with some inference. Eg., The woman raised her eyebrows in disapproval and pursed her lips angrily.
      \item Level 5: Complete inferences. Eg., The woman was furious, with angry eyes and pursed mouth.
  \end{enumerate}

\begin{figure*}[h]
  \centering
  \hspace*{-0.4cm}
  \includegraphics[scale=0.47]{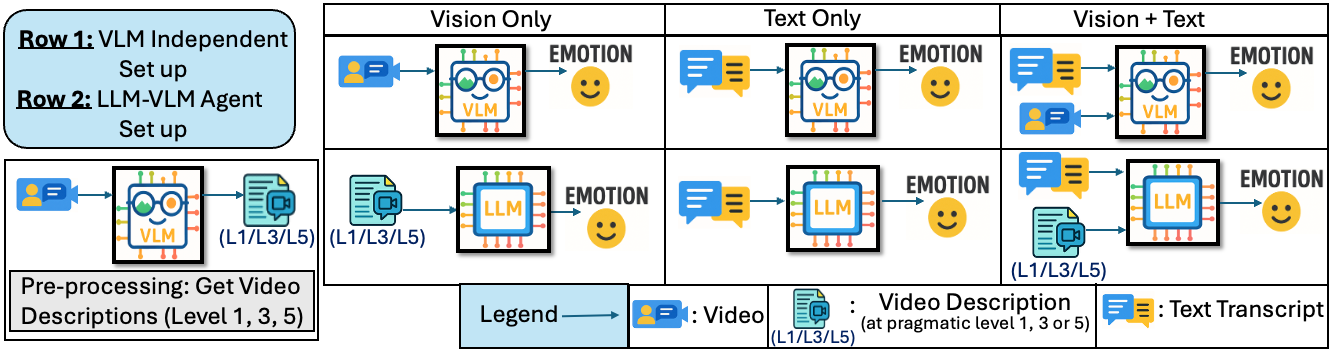}
  \caption{Emotion Prediction Experiment Set-ups. Legend on the bottom, preprocessing on the left.}
  \label{fig:vlm-settings}
\end{figure*}

\subsection{Diagnostic Emotion Prediction Task Overview}
The input for this task is the text-transcript of a conversation between multiple people and a video clip for the target utterance. The task is to predict the emotion of the speaker of the target utterance.

Several works such as \cite{Aviezer2012BodyCues} and \cite{barrett2010context} show that emotion recognition is not as simple as identifying facial expressions, but rather highly dependent on context. \cite{aviezer2011automaticity} showed that the same facial expression is interpreted differently by people when there are changes in body posture. Based on these works, in order for a VLM to do well at emotion prediction, it would need to be good at identifying relevant \textit{Visual Cues} in the video, but just as importantly, would need to excel at making the right \textit{VSP Inference} about the Visual Cues.

To the best of our knowledge, the capabilities of VLMs in this domain remain underexplored, and there are no datasets that isolate VSP Inference as a standalone task. For this reason, we chose emotion prediction as a starting point. This task served as a \textit{diagnostic tool} for us to first determine the capabilities of VLMs in social intelligence.

If a VLM struggles with the emotion prediction task, it could be due to hallucinations, or due to VSP inference mistakes (seeing the cues but misinterpreting them). The goal of the following proposed task was to investigate whether the VLM struggles with one or both types of mistakes. Our intent here was not to train the best multi-modal emotion predictor, but rather to gauge the social intelligence of a VLM. 

\subsection{Novel Visual Social-Pragmatic Inference Task Overview}
Visual Social-Pragmatic Inference is a new task that we propose in this paper. The input to the task is a video, a Visual Cue and two different VSP Inferences about the Visual Cue. The task is to pick the correct inference, based on visual context present in the video.

E.g., in Figure~\ref{fig:sad-howard}, the man is smiling. But in the video, he has tears in his eyes, and a slouched downcast body posture. His partner looks at him and rubs his hand sympathetically. Given the context, the correct interpretation is that the man is smiling sadly, reminiscing about something bittersweet.

With extensive human annotation, we constructed VIBE: a dataset of challenging VSP Inferences. VIBE was carefully curated to isolate the VSP Inference task. The Visual Cue that the VLM is required to interpret is guaranteed to be in the video, which mitigates hallucinations. Additionally, unlike other video datasets, the questions in VIBE are guaranteed to be answerable since we only have videos where the speaker is clearly visible, and the context required to interpret a Visual Cue is contained locally in the video.

Using VIBE, we conducted an analysis of several VLMs and were able to get a better understanding of VLMs abilities at VSP.


\section{Diagnostic Task: Emotion Prediction}
\label{sec:ep}
The conversations and video clips we used were sourced from the MC-EIU dataset \cite{mc-eiu} which consists of text transcripts of conversations and video clips of the utterances. Since our focus was videos that require VSP Inferences, we used only the videos explicitly labeled with a non neutral emotion. Additionally, it is known that VLMs struggle with long context videos \cite{qu2025doesvisionlanguagemodellost}, so we only used videos that were under 4 seconds. The final dataset included 3,536 Chinese video clips (from three different shows) and 5,589 English clips (from two shows). This formed a robust, diverse set of \textbf{over 9000 videos} spanning multiple \textbf{languages, cultures, and genres}. 

As illustrated in Figure~\ref{fig:vlm-settings}, we had two main diagnostic experimental settings: \textbf{(1) The VLM acts independently} to do the task, and \textbf{(2) The VLM acts as a perception agent} only, and a stronger LLM (GPT-4o-mini) reasons over the visual information. For both settings, we compare performance with the very strong baseline of an LLM (GPT-4o-mini) operating over the text transcript. The motivation behind having the two settings was to probe two different capabilities of the VLM. In setting (1) the goal was to test the VLM's ability to \textit{directly identify the emotion of the speaker}. It was not obvious how the VLM would perform compared to the stronger LLM; while the LLM has stronger reasoning capabilities, the VLM is aided by rich visual data in addition to the text. In contrast, it was reasonable to expect a boost in performance when the VLM had both modalities, compared to either one alone. In setting (2) the goal was to test the ability of the VLM to \textit{identify, accurately interpret and communicate} what it saw to the LLM (GPT-4o-mini). In this setting, it was reasonable to expect that the additional visual information would boost the performance of the LLM compared to the text only baseline. For both settings, to give the VLMs the best chance, we sampled the maximum possible frames that our compute resources allowed (30 frames). This decision was based on prior research that showed that performance generally increases as more frames are sampled \cite{frame-rate-1, frame-rate-2}.

\subsection{VLM Independent Setting}

We benchmarked the performance of 4 VLMs of sizes varying from 3B to 26B, from 2 different families of models: InternVL2 and Qwen \cite{chen2025expandingperformanceboundariesopensource, qwen2025qwen25technicalreport}. We chose the models based on their performance on vision benchmarks, support for video and the computational cost of running them. As illustrated in Figure~\ref{fig:vlm-settings}, we set up the emotion prediction task for each model under 3 different settings: (1)~Vision only, (2)~Text Only, and (3)~Text + Vision. We set the number of frames to be the maximum allowed frames for the smallest-context model (the InternVL2 models), which is 30 frames, and uniformly sampled this many frames for all models to ensure fairness. The 30-frame, 4-second limitations guaranteed at least 7 frames per second, but in most cases was more than 15.

\textbf{Diagnostic Results:} As seen in Table~\ref{tab:modality-performance}, the best performing model was InternVL2 26B with text only, having also very strong performance under the other two settings. Based on these results, we decided to use InternVL2 26B for further experiments. The key take-away here was, contrary to expectations, in general \textbf{\textit{combining the modalities did not perform significantly better}}, even though \textbf{\textit{each modality had strong individual performance}}.

\begin{table}[h]
\centering
\small
\begin{tabular}{lccc}
\toprule
\textbf{Model} & \textbf{Vision} & \textbf{Text} & \textbf{Text + Vision} \\
\midrule
GPT-4o-mini\footnotemark & --     & 0.538                & -- \\
\midrule
Qwen 3B                  & 0.387  & 0.446          & 0.456 \\
Qwen 7B                  & 0.373  & 0.449          & 0.476 \\
InternVL-8B              & 0.466  & 0.488          & 0.422 \\
InternVL-26B             & 0.457  & \textbf{0.493} & 0.468 \\
CogVLM2-Video            & 0.47   & --             & 0.387 \\
\bottomrule
\end{tabular}
\caption{Weighted F1 for VLM Independent Experiments.}
\label{tab:modality-performance}
\end{table}
\footnotetext{Strong closed model (skyline), to contextualize results.}

\subsection{VLM-LLM Agent Setting}
In this second diagnostic experiment, we used the VLM as a perception agent. We prompted the VLM to describe the speaker's facial expression and body language at 3 Levels of VSP Inference (see definitions) to get Video Descriptions and fed them into the LLM for classification. For some examples of the 3 Levels of VSP Inference, and an experiment confirming the toggling of levels see App. \ref{sec:app-a1}. In this experiment, we implemented the same three settings as in the previous experiment: V, T, and T + V (Figure~\ref{fig:vlm-settings}).

\begin{figure}[t]
  \centering
  \hspace*{-1cm}
  \includegraphics[scale=0.45]{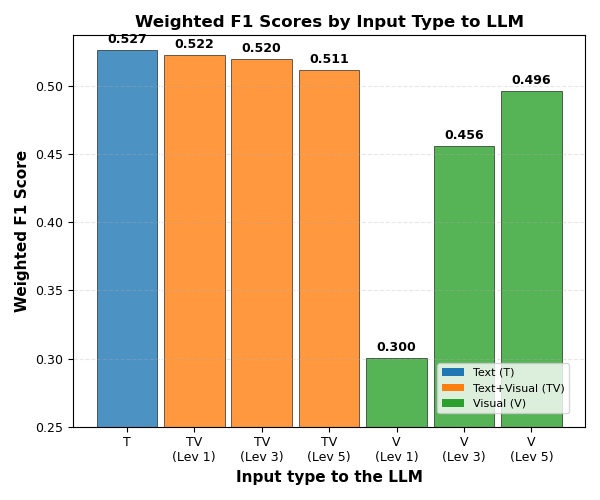}
  \caption{Weighted F1-Score for the VLM-LLM Agent experimental setting. The x-axis represents the various types of input to the LLM. Here Text is the text transcript of the conversation, and Visual means the visual cues generated from the VLM when given the conversation video. The Levels correspond to the various VSP levels of the visual cues (see definitions).}
  \label{fig:all_settings_weighted_f1}
\end{figure}

\textbf{Results:} This experiment's results are shown in Figure~\ref{fig:all_settings_weighted_f1}. The takeaways were: (1) There are strong signals in the vision which the VLM was able to communicate, (2) yet combining the modalities did not lead to significant improvement. Both the visual and textual modalities exhibited strong individual performance, but naively combining them did not yield a clear performance gain—when the VLM operated independently and also when paired with an LLM. Works like \cite{facial-mmt} suggest that directly incorporating vision is difficult because of the noise in the visual domain. To better understand the results, we broke them down by emotion. As shown in Figure~\ref{fig:weighted-f1-emotion} (App.~\ref{app:diagnostic}), the relative performance of each modality varies across emotions. We discuss the implications of these results in the next subsection.

\subsection{Emotion Prediction Discussion and Implications}

From the experiments in sections 4.1 and 4.2, we saw that while VLMs show reasonable strong performance over the visual domain, there was not a clear improvement over the text based performance (Table \ref{tab:modality-performance} and Figure \ref{fig:all_settings_weighted_f1}). This could imply two things: (1) That there is no visual information that would help improve over the text and/or (2) current VLMs do not have the ability to extract and communicate the necessary information accurately. We designed a simple yet effective algorithm to further analyze this: the Weighted Voting Algorithm. The success of the algorithm depended on two key conditions being true: (1) There is \textbf{complementary information} in the Video Descriptions that \textbf{cannot be recovered} from the text transcripts, and (2) The VLM's performance has a \textbf{consistent pattern}: it struggles with certain emotions and does better at others, even with small sample sizes. We first describe and present results of the algorithm, then discuss its implications:

\textbf{Weighted Voting Algorithm:} As shown in Figure~\ref{fig:all_settings_weighted_f1}, we had seven sources for emotion prediction: text-only (T), visual-descriptions at three pragmatic levels (V1, V3, V5), and combined modalities with the visual description at 3 pragmatic levels (TV1, TV3, TV5). We exclude Level 1 (due to subpar performance) from the weighted voting algorithm, and refer to the remaining 5 sources as \textit{Agents}. Each agent casts an equal vote, which is accepted if there is a clear majority. If no majority is reached, then the votes are weighted as follows:

For each of the five agents, we constructed a simple trust model by estimating their per-emotion precision on a small subset of the data, referred to as the \textit{Calibration Set}. Each agent casts a vote for an emotion, and the influence of their vote is weighted according to the corresponding trust score.

\begingroup
  \setlength{\abovedisplayskip}{4pt}
  \setlength{\belowdisplayskip}{4pt}
  \setlength{\abovedisplayshortskip}{3pt}
  \setlength{\belowdisplayshortskip}{3pt}

  \noindent
  Formally, let the agents be:
  \begin{equation}
    A = \{A_T, A_{V3}, A_{V5}, A_{TV3}, A_{TV5}\}
  \end{equation}

  \noindent
  Let the emotion space be:
  \begin{equation}
    E = \{\text{all candidate emotion labels}\}
  \end{equation}

  \noindent
   Using the calibration set, we compute the precision for each agent \(a\in A\) and emotion \(e\in E\) and use it to create a Trust function:
  \begin{equation}
    T: A \times E \rightarrow [0,1]
  \end{equation}

  \noindent
  Each agent \(a\in A\) votes for an emotion \(e_a\in E\). Their vote is weighted by the trust function T. Thus for each \(e\in E\) we compute a score, \(S(e)\):
  \begin{equation}
    S(e) = \sum_{a\in A} \mathbf{1}(e_a = e)\,T(a,e)
  \end{equation}

  \noindent
  Finally the predicted emotion \(\hat e\) is computed:
  \begin{equation}
    \hat e = \arg\max_{e\in E} S(e)
  \end{equation}
\endgroup


\underline{Overall Performance and Takeaways:} Results are in Table~\ref{tab:cal50-summary}. For both languages for both calibration settings we showed that our simple algorithm led to much better performance than any of the baselines. This validates our assumptions: \textit{(1) There is \textbf{complementary information} in video descriptions \textbf{not recoverable} from the text, and (2) The VLM struggles (and succeeds) in a \textbf{consistent} manner across emotions.} From these results, we can answer the following questions:
\begin{itemize}
    \item Is there complementary information in the videos beyond text? \(\rightarrow\) Yes
    \item Can a VLM extract and communicate this information? \(\rightarrow\) Yes (sometimes)
    \item Do VLM successes and failures follow systematic patterns? \(\rightarrow\) Yes
    \item Can these patterns be leveraged to do social and cognitive reasoning tasks better? \(\rightarrow\) Yes
\end{itemize}

Clearly, vision can play an important role in understanding social scenarios, and VLMs are able to (sometimes) give us valuable information that is not recoverable from the text. This encouraging result motivates us to further study exactly what kinds of things does a VLM struggle with? In the next section we show that hallucinations (well known in the literature) are not the only limitation: VLMs also struggle with VSPI. That is, even when they correctly identify visual elements, there are cases when they cannot interpret the correct VSP \emph{meaning} of what they identified.

\begin{table}[h]
\centering
\small
\begin{tabular}{lcccc}
\toprule
& \makecell[c]{Within\\lang en}
& \makecell[c]{Within\\lang zh}
& \makecell[c]{Cal. zh\\test en}
& \makecell[c]{Cal. en\\test zh} \\
\midrule
Vision       & 0.481                  & 0.526                  & 0.481                  & 0.526                  \\
Text         & 0.498                  & \underline{0.568}                 & 0.501                  & \underline{0.567}                  \\
Vis+Text     & \underline{0.509}                  & 0.543                  & \underline{0.509}                  & 0.538                  \\
Voting     & \textbf{0.523}
             & \textbf{0.592}
             & \textbf{0.530}
             & \textbf{0.609} \\
\bottomrule
\end{tabular}
\caption{Weighted Voting Algorithm at calibration size 50.}
\label{tab:cal50-summary}
\end{table}

\begin{figure*}[t]
  \centering
  \hspace*{-.2cm}
  \includegraphics[scale=0.45]{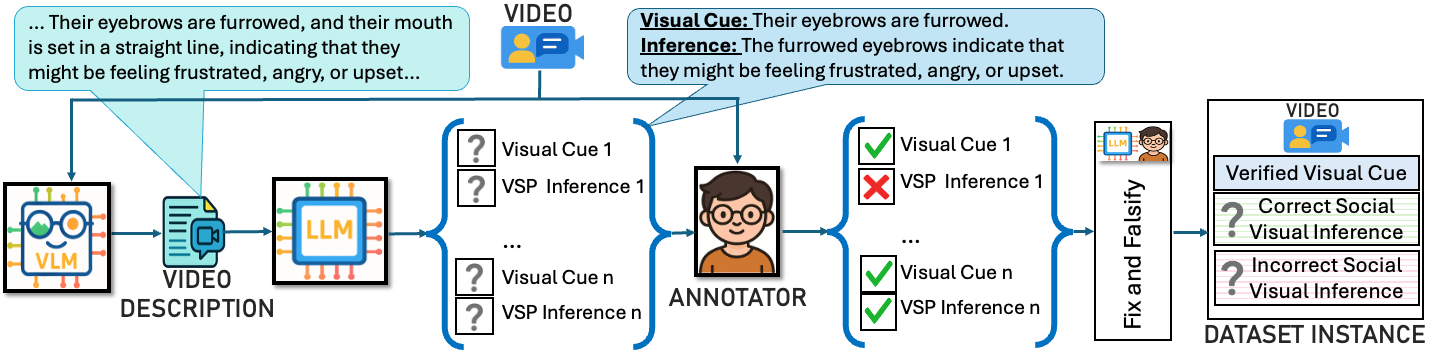}
  \caption{Main steps of the VIBE dataset curation pipeline.}
  \label{fig:annotation}
\end{figure*}

\section{Proposed Task: Visual Social-Pragmatic Inference}
The VSP Inference task is a new task we propose in this paper. Given a Visual Cue and a video, the task is to correctly interpret the pragmatic meaning of the cue. In the following sections we describe the process of creating the dataset and formalize the inputs and outputs. We benchmark select VLMs on the dataset and present the results and analysis.


\subsection{VIBE Dataset Creation}
This section details the process of curating the dataset we name VIBE (VSP Inference Based on Evidence). Our vision for VIBE was a dataset of rich video clips that contain Visual Cues which can be interpreted in more than one way. At a high level, the first step we took was to carefully filter for videos that were likely to contain such information. Once we had a pool of candidate videos, we began the process of curating the dataset. The main steps are shown in Figure~\ref{fig:annotation}, and detailed below.

\textbf{Video Selection:} To ensure the dataset consisted of videos rich in challenging Visual Cues, we used performance on the emotion prediction task as a heuristic for video selection. We note that the emotion prediction task was originally designed for a multi-modal setting. Whereas for our dataset, we wanted to isolate videos that were informative and interpretable based on visual information \emph{alone}, as is required by the VSP Inference task. To that end, we applied a three-stage filtering process to identify promising videos. Specifically, we selected videos where the emotion was misclassified for three different settings: (1) VLM given only the video, (2) LLM provided with Level 3 visual information, and (3) LLM provided with Level 5 visual cues. Furthermore, since some emotion pairs naturally co-occur very frequently (e.g., joy and surprise), we only retained videos where the predicted and gold emotions are highly unlikely to co-exist, such as joy and anger. The list of excluded and retained emotion pairs is provided in App.~\ref{app:implementation}.

\textbf{Video Description Generation:} For each video, in our candidate pool of videos, we extracted Video Descriptions that would be pertinent to the dataset. We generated these descriptions to be at Level 3 and Level 5 of VSP Inference and made sure there was a focus on facial expressions and body language. The prompts can be found in App.~\ref{app:prompts}.

\textbf{Visual Cue and Inference Extraction:} We implemented few-shot prompting to GPT-4o to break down the descriptions into candidate\footnote{"candidate" since Visual Cues are unverified at this stage.} Visual Cues and VSP Inferences.

\textbf{Human Annotation: } A human was shown the video clip and asked to predict the main speaker's emotion. The annotator could pick up to 3 Ekman emotions. Next, they verified the presence (or absence) of the candidate Visual Cues in the video. If the Visual Cue was confirmed to be present in the video, they rated the correctness of the Pragmatic Visual Inference drawn from the Visual Cue.

For every Visual Cue-VSP Inference pair, we had 2-4 humans do the annotation. We chose annotators from a variety of backgrounds and included both native English speakers and native Chinese speakers. All annotators had at least a Bachelor's degree and were proficient in English. The annotation involved a total of about 25 hours of high quality human annotation time. Even though this was an inherently subjective task, we saw an agreement of 74.5\% for visual cues (random being 50\%), and 52.8\% for VSP inferences (random being 33\%). To ensure the high quality of VIBE, the dataset only includes Visual Cue-VSP Inference pairs for which at least 2 annotators had perfect agreement on both Visual Cue and VSP inference.

\textbf{Fix and Falsify: } Post human annotation, we were left with a set of verified Visual Cues, VSP Inferences that were confirmed to be correct or incorrect, and a pool of emotions for the speaker labeled by annotators. For each video, we computed a "max scoring emotion", which was the emotion that was labeled the most times by annotators (ties broken arbitrarily). We envisioned VIBE as a multiple choice dataset, so as the final step, we needed to come up with correct and incorrect counter-parts for all the inferences we had. We used an LLM (GPT-4o-mini) and the human labeled emotions to do this final step of "fixing" and "falsifying" the inferences.\\
\underline{Creating Correct Inference Choices:} For all the inferences that were confirmed \textit{incorrect} by annotators, we came up with the correct inference using the following methods:
\begin{enumerate} [itemsep=0pt, topsep=0pt]
    \item When the max scoring emotion from annotators agreed with the gold label for the video, we used this emotion to correct the wrong inference. We call these \textit{human fixed inferences}.
    \item When the agreement was not there, we simply negated the incorrect inference. We call these \textit{fixed by negation inferences}.
\end{enumerate}
\underline{Creating Incorrect Inference Choices:} For all the inferences that were confirmed \textit{correct} by annotators, we came up with the incorrect inference using the following methods:
\begin{enumerate} [itemsep=0pt, topsep=0pt]
    \item When the max scoring emotion from annotators agreed with the gold label for the video, we used this emotion to come up with a mutually exclusive emotion (App.~\ref{app:implementation}). We then used this emotion to falsify the inference. We call these \textit{distractor inferences}.
    \item When an agreement could not be reached, we employed two methods: We either used the emotion that no annotator voted for to falsify the inference (these are also distractor inferences) or we simply negated the correct inference, which we call \textit{false negation inferences}.
\end{enumerate} 
\underline{Final Dataset Composition:} In the final dataset we had 50\% negation style choices (which could be correct or incorrect), and 50\% distractor style choices (which could also be correct or incorrect). The diverse methods were carefully designed to safeguard the dataset against being hacked by always choosing negations or distractors or even by discerning between VLM and LLM generated text.

\textbf{Final Dataset: } The final dataset contains 433 unique video clips and 994 unique instances of the VSP Inference task. An instance of the task is: Given a video, a Visual Cue, and two candidate VSP Inferences, output the correct inference.

\subsection{VSP Inference Experiments}
We benchmarked all VLMs used in the emotion prediction task on our VIBE dataset to illustrate its difficulty. We additionally include the expensive OpenAI GPT-4o mini model as a reference to large, closed VLMs. All models are prompted using CoT, though we include results for standard prompting in App.~\ref{app:benchmark-vibe}. Finally, we had two humans do the VIBE dataset on 100 instances each as comparison. The results can be seen in Table~\ref{tab:accuracy-cot}. These results demonstrate that  (1) VLMs struggle with VSPI (which comes intuitively to humans), and (2) The question types in VIBE that VLMs struggle with most (fixed and NT) are consistently hard for all the models we evaluated (but not for humans!). This result underscores the importance of targeting VSPI for VLMs in order to make them socially and emotionally intelligent. In section 5.3 we dive into a deeper analysis of the kinds of mistakes made on VIBE and connect the results back to the emotion prediction task from Section 4

\begin{table}[h]
\centering
\footnotesize
\begin{tabular}{lcccc|c}
\toprule
\textbf{Model} & \textbf{Fixed} & \textbf{Distractor} & \textbf{NF} & \textbf{NT} & \textbf{All} \\
\midrule
IVL2-26B\footnotemark                 & 29.5   & 74.5               & 78.8               & 18.1                & 63.4 \\
IVL2-8B                   & 13.1               & \underline{93.5}      & \textbf{97.3}      &  5.5                & 73.8 \\
IVL2-4B                   & 20.5               & 86.6               & 87.7               & 29.1       & 71.5 \\
\midrule
Q2.5-3B                   & 25.4               & 90.3               & 93.8               & 22.8                & 75.1 \\
Q2.5-7B                   & 26.2               & 93.3   & 87.7               & \underline{26.8}    & 74.4 \\
\midrule
4o-mini                   & \underline{31.1}     & 90.1               & 94.6   & 17.3                & \underline{75.3} \\
\midrule
CogVLM\\2-Video             & 25.4               & 78.0               & 95.4               & 12.6                & 69.7 \\
\midrule
Humans           & \textbf{88.9}           & \textbf{94.7}          & \underline{95.2}           & \textbf{81.8}            & \textbf{92.5} \\
\bottomrule
\end{tabular}
\caption{Accuracy on VIBE by model and question type (CoT). NF and NT are \emph{negation false} and \emph{negation true} respectively.}
\label{tab:accuracy-cot}
\end{table}
\footnotetext{Evaluation on this model is not wholly fair as the dataset was partially created from its mistakes.}

\subsection{VSP Inference Analysis}
\label{sec:vsp-analysis}
In this section, we use the VIBE dataset to quantitatively analyze the VLM's ability to describe a scene (i.e. \emph{read the room}). We do this by investigating the impacts of its visual cues on the downstream emotion prediction task. We aim to answer two questions: (1) ``Which kinds of visual cues help reach the correct emotion label?'' and (2) ``When a mistake is made, is it due to hallucination, misinterpretation, or something else (like an uninformative visual cue, necessary textual context, etc.)?''.


From our annotations, using only annotator-agreed labels, we had over 1.2 K visual cues. Among these, 27.3\% were labeled as hallucinations. In the remaining, non-hallucinated visual cues, 27.5\% were labeled as being misinterpreted.

\begin{figure}[t]
  \centering
  \includegraphics[width=\linewidth]{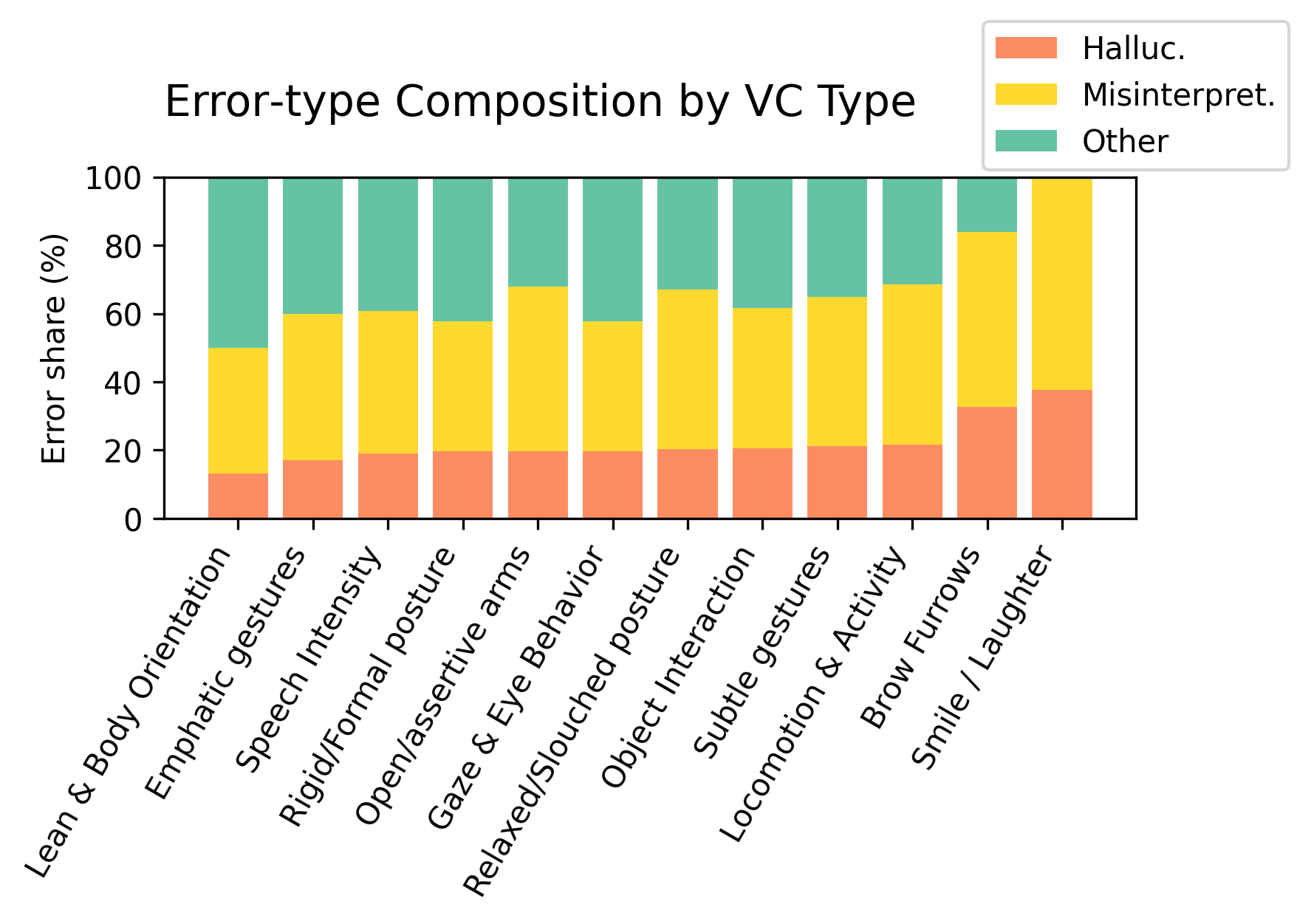}
  \caption{Distribution of error type by type of visual cue.}
  \label{fig:error-type-compo}
\end{figure}

In order to better understand the distribution, we came up with 12 clusters of visual cues (e.g. ``Emphatic gestures'', ``Smile/Laughter'', etc.), created keyword representations of them, and used cosine similarity over SBERT embeddings to retrieve the top 200 visual cues for each. Since all visual cues in the dataset were there because they led to an incorrect emotion prediction, we aimed to explain that error using the human annotations. Specifically, we calculated the hallucination and misinterpretation rate within each cluster. If the visual cue was neither hallucinated nor misinterpreted, then it was correct, but for other reasons led to an incorrect emotion prediction (e.g. being uninformative, textual context being necessary, etc.). Figure~\ref{fig:error-type-compo} shows the composition of the kinds of errors among the different visual cue clusters.

We see that the ``Smile/Laughter'' cluster has both a high misinterpretation and hallucination rate, meaning that the VLM often mistakes non-smiles for smiles, and even when it identifies a real smile, it often cannot interpret its meaning (e.g. a \emph{sad smile}). Likewise, we see that the cluster ``Leaning \& Body Orientation'' visual cues (e.g. ``leaning forward'', ``upright posture'', ``laying down'') are less commonly hallucinated, but often misinterpreted.

\begin{figure}[ht]
  \centering
  \includegraphics[width=\linewidth]{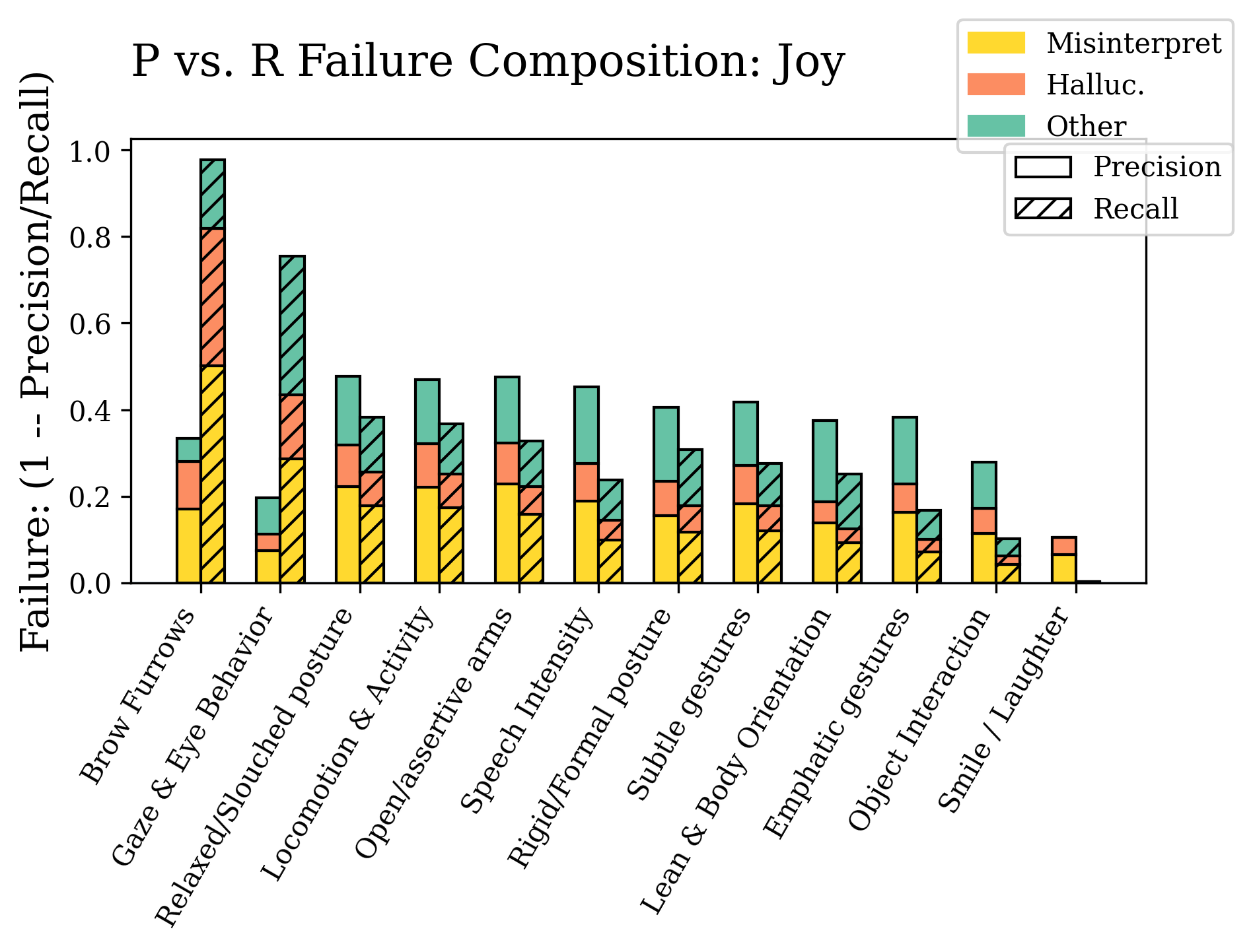}
  \caption{Error analysis for `Joy'. Bigger bars = bigger errors.}
  \label{fig:p-r-joy}
\end{figure}

In order to further investigate how these errors led to misclassification, we mapped the clusters onto the full, un-annotated space of visual cues used for the emotion prediction task. To do this, we used the same cluster keywords and the same cosine similarity search mechanism to retrieve the top 500 visual cues for each cluster. We then computed the precision and recall for each emotion, for each cluster. The result for `Joy' can be seen in Figure~\ref{fig:p-r-joy} (other emotions in App.~\ref{app:failure-analysis}). We plot $1 - P$ and $1 - R$ on the y-axis to emphasize bigger errors and visualize their compositions.

We see that for `Joy', recall failures substantially exceed precision failures across every cluster, indicating that the VLM more often \emph{misses true visual cues} than it hallucinates spurious ones. Particularly, subtle facial movements such as ``Brow Furrows'' and ``Gaze \& Eye Behavior'' exhibit the highest total failure rates, with nearly all true instances going undetected (very low recall). In sharp contrast, the `Smile/Laughter'' cluster is almost perfectly handled (both precision and recall failures < 0.1), showing the model’s strength on overt expressions like smiles. Across the board, misinterpretation comprises the largest slice of the error budget, followed by hallucination, while other error types remain minimal. These results suggest that, to improve emotion‐prediction accuracy, future work should focus on bolstering the model’s sensitivity to subtle nonverbal signals and reducing its tendency to misinterpret correctly detected cues.

\section{Conclusion}
We introduced the VSP Inference task, exposing a previously overlooked limitation in VLMs. Through diagnostic analyses, human annotation and the VIBE dataset, we showed that current VLMs struggle with VSP Inference. Our results underscore the need for improved social reasoning in VLMs and provide a benchmark for future research.

\section{Limitations and Future Work}

\textbf{Variety in Vision models chosen: } In this paper we limited our analyses to 3 families of VLMs up to size 26B. The VIBE dataset was tested on 7 models of varying families and sizes. More extensive analysis including more models and architectures could lead to more insights.

\textbf{Bias Due to Video Description Source: } The Visual Cues in our dataset were selected from failure cases of InternVL2 26B, which may have inherent biases. Anything the model did not identify as relevant would be missed in the dataset. The dataset would benefit from having more models in the video description generation process.

\textbf{Inherent subjectivity in human annotation: } While we saw high inter-annotator agreement, the annotation task was inherently subjective. Cultural and personal biases may have some impact on the dataset, even if it only meant that we were not able to include some hard cases because of disagreement within annotators.

\textbf{Future Work}
The limitations mentioned above give rise to some natural directions for future work. It would be worthwhile to study more models on this family of tasks, with an emphasis on architecture and training data to see if there are significant performance differences. Additionally, given the subjective nature of the task, a promising extension of this paper would be to model variations in human judgments as uncertainty estimation of the models. Finally, this paper is a first step into investigating the alignment of multi-modal models and human perceptions of social scenarios; We believe that further investigations and subsequent improvements to these models will lead to socially competent AI systems.

\section{Ethics Statement}
This paper addresses social reasoning capabilities of VLMs, and proposes a dataset for it. We acknowledge that the inherent subjective nature of the task might affect the dataset. To mitigate any issues we employed a diverse group of annotators that were willing to volunteer their time. We recognize that systems that aim to mimic or understand human social dynamics raise ethical concerns. The purpose of this paper is not to endow VLMs with any malicious capabilities but rather to foster better understanding of such models in the research community. The video clips in our data-set were sourced from an existing research dataset and were obtained with the required permissions. This data is meant to be used for research purposes only.

\section*{Acknowledgments}
We thank the reviewers for their insightful comments that helped improve the paper. This work
was supported by NSF CAREER award IIS2048001 and the DARPA CCU program. The contents are those of the author(s) and do not necessarily represent the official views of, nor an endorsement by, DARPA, or the US Government.

\bibliography{custom}

\appendix

\section{Diagnostic Results}
\label{app:diagnostic}

\subsection{Experiment to support toggling Levels of VSP Inference}
\label{sec:app-a1}
In order to generate output at the correct level of VSP Inference, prompts were manually tuned with care. Additionally we did an experiment to confirm that the outputs generally matched the expected VSP Inference Level; We extracted the top 20 n-grams (n=5, 7, 10) from each of the 3 levels. These were manually inspected (and 10 are provided below in Table\ref{tab:ngrams}. Additionally, the 20 10-grams were provided to GPt o4-mini-high to compare them with shuffled level order using the prompt:

\begin{figure}[h!]
    \centering
    \begin{tcolorbox}[colback=gray!30, colframe=black, boxrule=0.5mm, width=\columnwidth]
    \tiny
    I’m going to give you top phrases from three different sources. Summarize the main differences between the 3 sources. Output a table where each row is a source. The columns should be “Focus” and “Characteristic Phrases”.\\\\
    Source x: <top n-grams from level 1>\\
    Source y: <top n-grams from level 5>\\
    Source z: <top n-grams from level 3>\\
    \end{tcolorbox}
    \caption{Prompt to confirm VSP Level.}
    \label{fig:rebuttal-exp}
\end{figure}

The summary result (re-ordered for readability):
\begin{table}[h]
\centering
\footnotesize
\begin{tabular}{p{0.95\linewidth}}
\toprule
\textbf{Level 1 (Source: X)} \\
\midrule
\textbf{Focus:} Detailed, low-level facial movements \\
\textbf{Characteristic Phrases:} \\
-- ``his eyebrows move up and down'' \\
-- ``her mouth and lips open and close as she speaks'' \\
\addlinespace[0.6em]
\textbf{Level 3 (Source: Z)} \\
\midrule
\textbf{Focus:} Broader body-language/facial cues \\
\textbf{Characteristic Phrases:} \\
-- ``exhibits a range of body language cues that suggest she is'' \\
-- ``his eyebrows are slightly furrowed indicating concern'' \\
\addlinespace[0.6em]
\textbf{Level 5 (Source: Y)} \\
\midrule
\textbf{Focus:} Explicit emotion labeling/classification \\
\textbf{Characteristic Phrases:} \\
-- ``emotions such as joy sadness fear disgust surprise or anger'' \\
-- ``he is feeling joy. he is smiling and appears to be'' \\
\bottomrule
\end{tabular}
\caption{The Focus and Characteristic Phrases confirm that the toggling worked as expected.}
\label{tab:levels_sources_single}
\end{table}

\begin{table*}[t]
\centering
\scriptsize
\resizebox{\textwidth}{!}{%
\begin{tabular}{p{0.28\linewidth}r p{0.28\linewidth}r p{0.28\linewidth}r}
\toprule
\textbf{10-gram} & \textbf{Freq} & \textbf{7-gram} & \textbf{Freq} & \textbf{5-gram} & \textbf{Freq} \\
\midrule
\multicolumn{6}{l}{\textbf{Level 1 (Detailed, low-level facial movements)}} \\
changes in his facial features the video. his eyebrows move & 1031 & mouth and lips open and close as & 2117 & eyebrows move up and down & 3621 \\
and her mouth and lips open and close as she & 1020 & maintains a relatively static posture the video. & 1975 & the the man in the & 2690 \\
her mouth and lips open and close as she speaks. & 1005 & changes in his facial features the video. & 1294 & mouth and lips open and & 2482 \\
changes in her facial features the video. her eyebrows move & 998 & in his facial features the video. his & 1294 & and lips open and close & 2473 \\
facial features the video. his eyebrows move up and down & 975 & his facial features the video. his eyebrows & 1294 & there are no significant changes & 2433 \\
in his facial features the video. his eyebrows move up & 960 & changes in her facial features the video. & 1180 & are no significant changes in & 2433 \\
his facial features the video. his eyebrows move up and & 960 & in her facial features the video. her & 1178 & lips open and close as & 2128 \\
and his mouth and lips open and close as he & 855 & her facial features the video. her eyebrows & 1178 & maintains a relatively static posture & 2040 \\
in her facial features the video. her eyebrows move up & 854 & eyebrows move up and down her forehead & 1144 & a relatively static posture the & 2007 \\
her facial features the video. her eyebrows move up and & 854 & her mouth and lips open and close & 1142 & relatively static posture the video. & 1975 \\
\quad ... & & \quad ... & & \quad ... & \\
\midrule
\multicolumn{6}{l}{\textbf{Level 3 (Broader body-language/facial cues)}} \\
a range of body language cues that suggest she is & 647 & exhibits a range of facial expressions the & 2150 & a range of facial expressions & 3760 \\
exhibits a range of body language cues that suggest she & 644 & a range of facial expressions the video. & 1798 & exhibits a range of facial & 3704 \\
exhibits a range of facial expressions that suggest he is & 601 & exhibits a range of facial expressions that & 1522 & is feeling a mix of & 3547 \\
that suggest she is feeling a mix of emotions. her & 544 & exhibits a range of body language cues & 1488 & the a man in a & 2297 \\
exhibits a range of body language cues that suggest he & 513 & a range of body language cues that & 1482 & he is engaged in a & 2285 \\
a range of body language cues that suggest he is & 513 & range of body language cues that suggest & 1430 & range of facial expressions the & 2192 \\
that suggest he is feeling a mix of emotions. his & 512 & a range of facial expressions that suggest & 1332 & is engaged in a serious & 2036 \\
his eyebrows are slightly furrowed indicating concentration or concern. his & 501 & his body language suggests that he is & 1156 & body language cues that suggest & 1941 \\
eyebrows are slightly furrowed indicating concentration or concern. his mouth & 482 & shirt exhibits a range of facial expressions & 938 & be feeling a mix of & 1891 \\
are slightly furrowed indicating concentration or concern. his mouth is & 482 & the slight furrow in his brow and & 936 & that he might be feeling & 1831 \\
\quad ... & & \quad ... & & \quad ... & \\
\midrule
\multicolumn{6}{l}{\textbf{Level 5 (Explicit emotion labeling)}} \\
emotions such as joy sadness fear disgust surprise or anger. & 680 & the in the appears to be feeling & 2188 & the in the is the & 5982 \\
the in the appears to be expressing a mix of & 510 & the in the appears to be expressing & 1446 & in the appears to be & 4947 \\
strong emotions such as joy sadness fear disgust surprise or & 440 & joy sadness fear disgust surprise or anger. & 1345 & the in the appears to & 4808 \\
the in the appears to be expressing a sense of & 406 & appears to be engaged in a conversation & 1335 & to be engaged in a & 3396 \\
such as joy sadness fear disgust surprise or anger. based & 404 & in the is the man in the & 1247 & in the is the man & 3015 \\
as joy sadness fear disgust surprise or anger. based on & 404 & the in the is the man in & 1244 & the appears to be feeling & 2554 \\
that he is feeling joy. he is smiling and appears & 383 & appears to be expressing a mix of & 1226 & appears to be engaged in & 2534 \\
based on these observations it seems that the is feeling & 380 & to be engaged in a conversation with & 1054 & in the is the woman & 2104 \\
a positive and happy emotion. the in the is the & 349 & in the appears to be expressing a & 963 & appears to be expressing a & 2090 \\
is feeling joy. he is smiling and appears to be & 329 & the in the is the man wearing & 946 & her body language including her & 1685 \\
\quad ... & & \quad ... & & \quad ... & \\
\bottomrule
\end{tabular}}
\caption{Representative n-grams (10, 7, and 5) across Levels 1, 3, and 5.}
\label{tab:ngrams}
\end{table*}

\subsection{Complete VLM-LLM Agent Experiment Results}

Here are the f1-scores of the VLM-LLM agent setting for both languages (and overall). Chinese scores are generally slightly higher than the Egnlish ones, which could be due to quirks of the VLM model or the data itself.
\begin{table}[h]
\centering
\small
\begin{tabular}{lccc}
\toprule
\textbf{Modality} & \textbf{Overall} & \textbf{Chinese} & \textbf{English} \\
\midrule
T     & 0.5266 & 0.5673 & 0.5006 \\
TV\_1 & 0.5224 & 0.5582 & 0.4991 \\
TV\_3 & 0.5199 & 0.5376 & 0.5091 \\
TV\_5 & 0.5114 & 0.5422 & 0.4952 \\
V\_1  & 0.3002 & 0.2326 & 0.3438 \\
V\_3  & 0.4559 & 0.4568 & 0.4583 \\
V\_5  & 0.4961 & 0.5260 & 0.4805 \\
\bottomrule
\end{tabular}
\caption{Weighted F1-scores for LLM operating over various modalities. Visual Cues are generated by a VLM given videos of the conversation.\\T: Transcript, TV: Transcript \& Visual Cues, V: Visual Cues}
\label{tab:llm-agent-f1-scores}
\end{table}

\subsection{Weighted Voting Algorithm Details and Full Results}

We split our dataset into a very small calibration pool (300 chinese, 550 english) which still left a large test set to test on (3236 for chinese, 5039 for english). For various calibration sizes, we randomly sampled from the calibration pool, and repeated the algorithm described above 50 times. As shown in Table~\ref{tab:combined-calibration} even at just calibration size of 50, we were able to do much better than any of the other LLM classifications. We also report standard error and confidence intervals for the calibration sets.

\begin{table}[h]
\centering
\scriptsize
\begin{tabular}{lcccc}
\toprule
\textbf{Cal size} & \textbf{Text} & \textbf{Vision} & \textbf{Vision+Text} & \textbf{Ensemble} \\
\midrule
\multicolumn{5}{l}{\textbf{Within Language: en (Test: 5039, Calibration Pool: 550)}} \\
0   & 0.498 & 0.481 & 0.509 & 0.522 \\
50  & 0.498 & 0.481 & 0.509 & 0.523 (7.49e-04, 1.51e-03) \\
100 & 0.498 & 0.481 & 0.509 & 0.523 (6.88e-04, 1.38e-03) \\
250 & 0.498 & 0.481 & 0.509 & 0.524 (5.52e-04, 1.11e-03) \\
500 & 0.498 & 0.481 & 0.509 & 0.526 (1.45e-04, 2.91e-04) \\
\midrule
\multicolumn{5}{l}{\textbf{Within Language: zh (Test: 3236, Calibration Pool: 300)}} \\
0   & 0.568 & 0.526 & 0.543 & 0.567 \\
50  & 0.568 & 0.526 & 0.543 & 0.592 (2.42e-04, 4.85e-04) \\
75  & 0.568 & 0.526 & 0.543 & 0.592 (1.72e-04, 3.46e-04) \\
100 & 0.568 & 0.526 & 0.543 & 0.592 (1.16e-04, 2.33e-04) \\
250 & 0.568 & 0.526 & 0.543 & 0.592 (6.17e-05, 1.24e-04) \\
\midrule
\multicolumn{5}{l}{\textbf{Cross Language: zh Calibration, en Test (Test: 5589, Cal Pool: 300)}} \\
0   & 0.501 & 0.481 & 0.509 & 0.542 \\
50  & 0.501 & 0.481 & 0.509 & 0.530 (5.10e-04, 1.02e-03) \\
75  & 0.501 & 0.481 & 0.509 & 0.530 (3.97e-04, 7.98e-04) \\
100 & 0.501 & 0.481 & 0.509 & 0.530 (3.42e-04, 6.87e-04) \\
250 & 0.501 & 0.481 & 0.509 & 0.531 (1.56e-04, 3.13e-04) \\
\midrule
\multicolumn{5}{l}{\textbf{Cross Language: en Calibration, zh Test (Test: 3536, Cal Pool: 550)}} \\
0   & 0.567 & 0.526 & 0.538 & 0.598 \\
50  & 0.567 & 0.526 & 0.538 & 0.609 (1.65e-03, 3.33e-03) \\
100 & 0.567 & 0.526 & 0.538 & 0.613 (1.16e-03, 2.33e-03) \\
250 & 0.567 & 0.526 & 0.538 & 0.614 (5.87e-04, 1.18e-03) \\
500 & 0.567 & 0.526 & 0.538 & 0.614 (2.97e-04, 5.97e-04) \\
\bottomrule
\end{tabular}
\caption{Results from Weighted Voting Algorithm}

\label{tab:combined-calibration}
\end{table}

\begin{figure*}[ht]
  \centering
  \hspace*{-1cm}
  \includegraphics[scale=0.6]{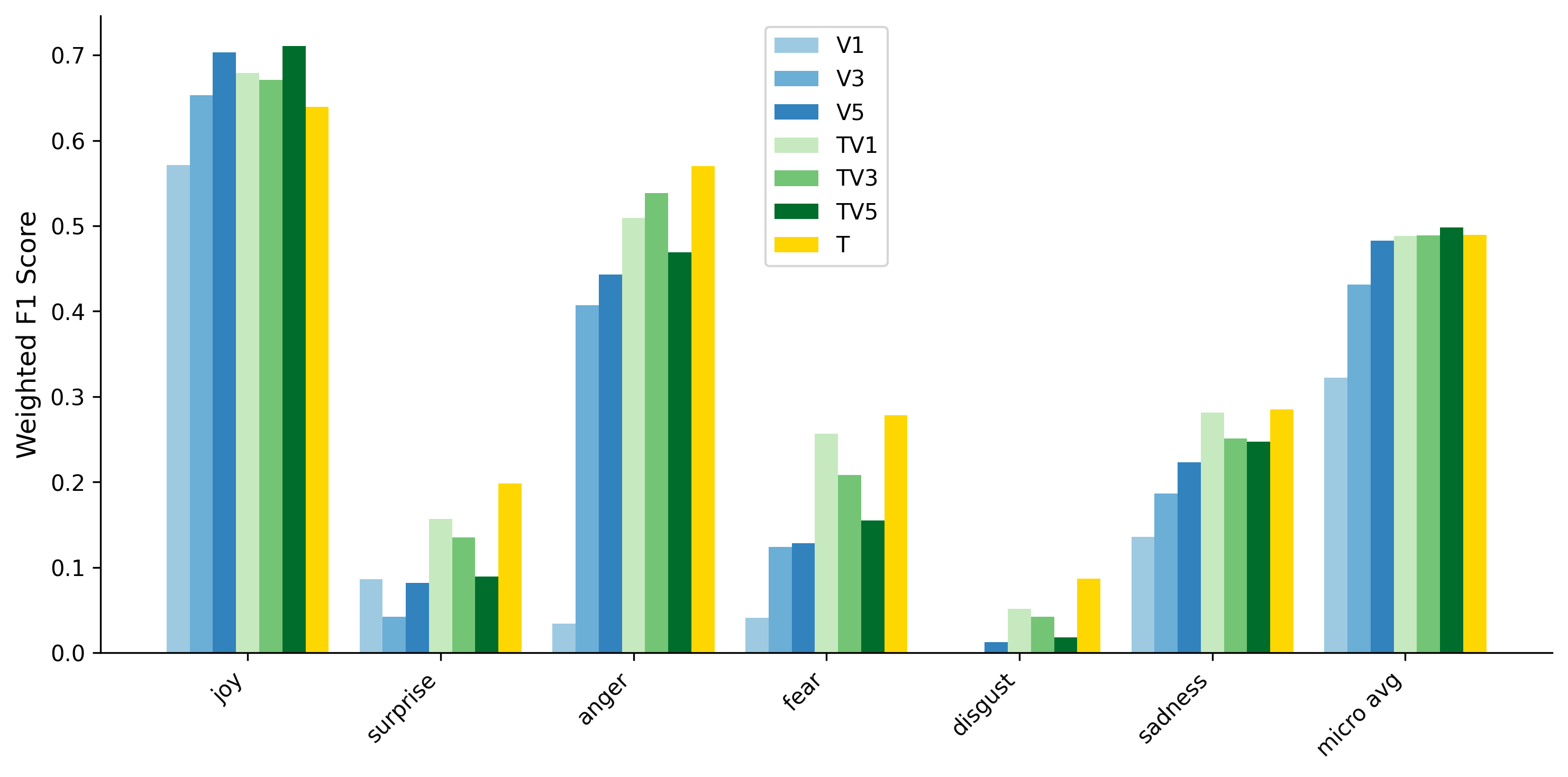}
  \caption{Weighted f1-by emotion.}
  \label{fig:weighted-f1-emotion}
\end{figure*}

\section{Benchmarking on VIBE}
\label{app:benchmark-vibe}
Table~\ref{tab:accuracy-direct} shows results for benchmarking on VIBE using a direct prompt instead of CoT prompting.

\begin{table}[h]
\centering
\scriptsize
\begin{tabular}{lcccc|c}
\toprule
\textbf{Model} & \textbf{Fixed} & \textbf{Distractor} & \textbf{NF} & \textbf{NT} & \textbf{All} \\
\midrule
IVL2-26B                  & 24.6 & 88.7 & 93.8 & 19.7 & 73.9 \\
IVL2-8B                   & 31.1 & 87.1 & 94.4 & 15.0 & 73.7 \\
IVL2-4B                   & 27.9 & 88.2 & 90.6 & 29.1 & 74.1 \\
\midrule
Q2.5-3B                   & 30.3 & 86.9 & 92.3 & 16.0 & 72.8 \\
Q2.5-7B                   & 26.2 & 88.2 & 93.0 & 19.7 & 73.6 \\
\bottomrule
\end{tabular}
\caption{Accuracy by model and question type (Direct Prompt).}
\label{tab:accuracy-direct}
\end{table}

\begin{figure}[h!]
    \centering
    \begin{tcolorbox}[colback=gray!30, colframe=black, boxrule=0.5mm, width=\columnwidth]
    \tiny
    \#\#\# Task Overview\\
    You are a socially intelligent body language expert. Your task is to interpret a person's body language. You will be given a video clip with one main speaker and asked which interpretation of their body language is better. Follow the Task Guidelines and the Response Format.\\
    
    \#\#\# Task Guidelines\\
    - You are given a video clip with one main speaker.\\
    - You are given one fact about the speaker's body language, and two possible interpretations of that body language.\\
    - Think out loud about which interpretation is better given what you see in the video (2-3 sentences).\\
    - Finally, give your answer according to the Response Format.\\
    
    \#\#\# Response Format\\
    Thinking out loud: <your thoughts about which interpretation is better (2-3 sentences)>\\
    Answer: <A OR B>\\
    
    \#\#\# Video Clip\\
    \{clip\}
    \#\#\# Fact \\
    \{ fact\_text \} \\
    
    \#\#\# Interpretations\\
    A. \{ inference\_A \} \\ 
    B. \{ inference\_B \} \\
    
    \#\#\# Response
    \end{tcolorbox}
    \caption{CoT prompt used for all VLMs on the VIBE benchmarking.}
    \label{fig:vlm-f-3}
\end{figure}

\section{Failure Analysis}
\label{app:failure-analysis}
\begin{figure}[ht]
  \centering
  \includegraphics[width=\linewidth]{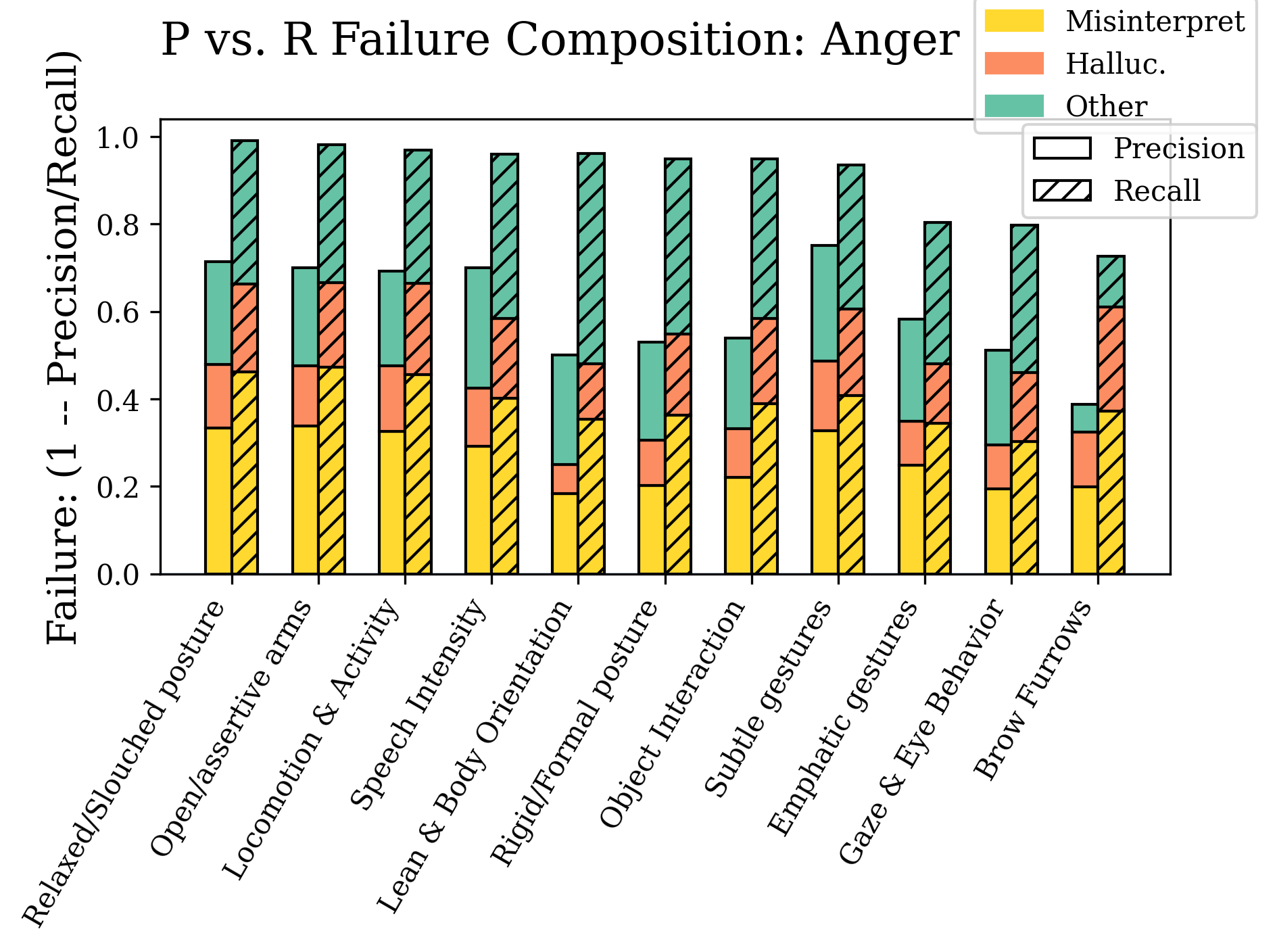}
  \caption{Error analysis for `Anger'. Bigger bars = bigger errors.}
  \label{fig:p-r-anger}
\end{figure}

\begin{figure}[ht]
  \centering
  \includegraphics[width=\linewidth]{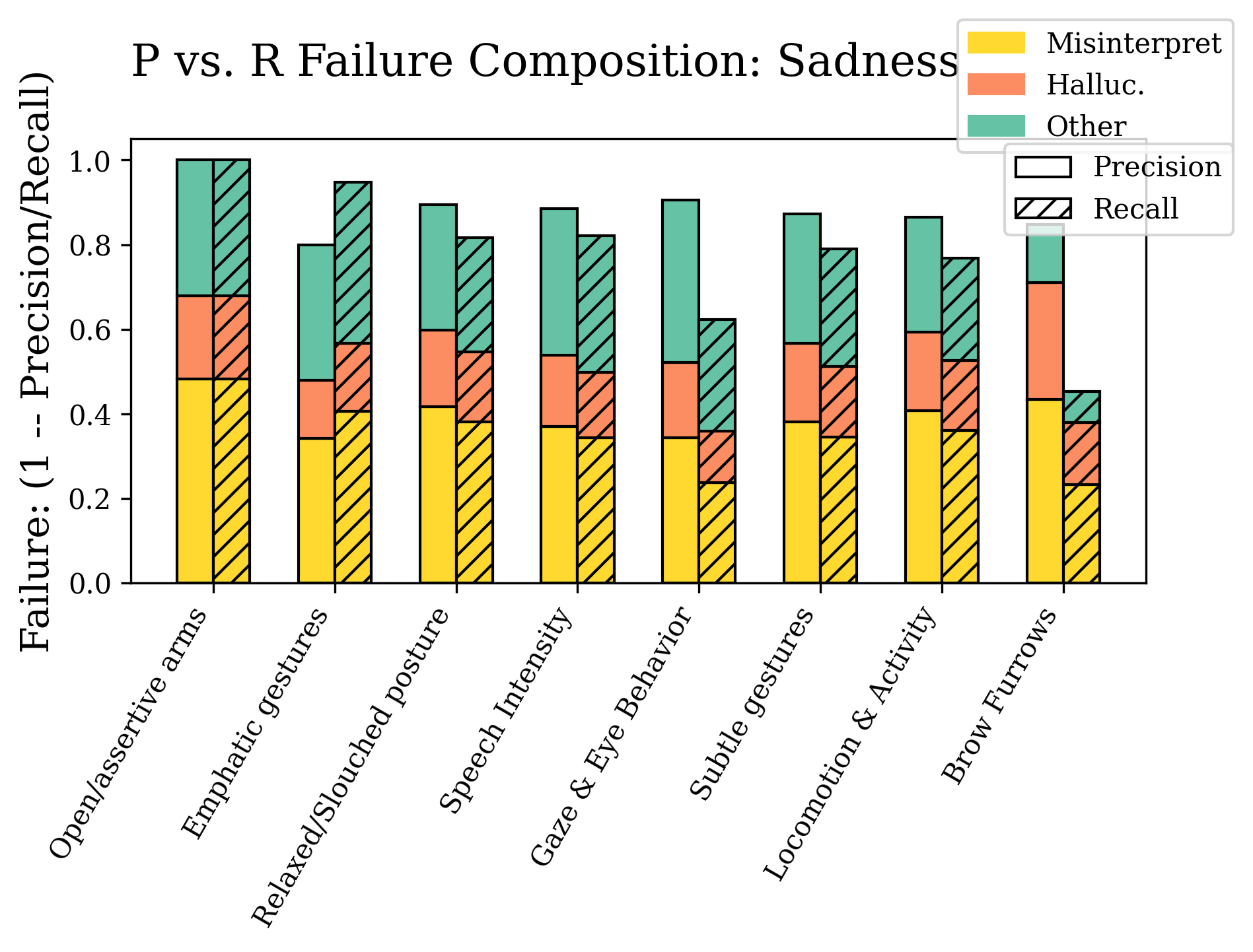}
  \caption{Error analysis for `Sadness'. Bigger bars = bigger errors.}
  \label{fig:p-r-sadness}
\end{figure}

\begin{figure}[ht]
  \centering
  \includegraphics[width=\linewidth]{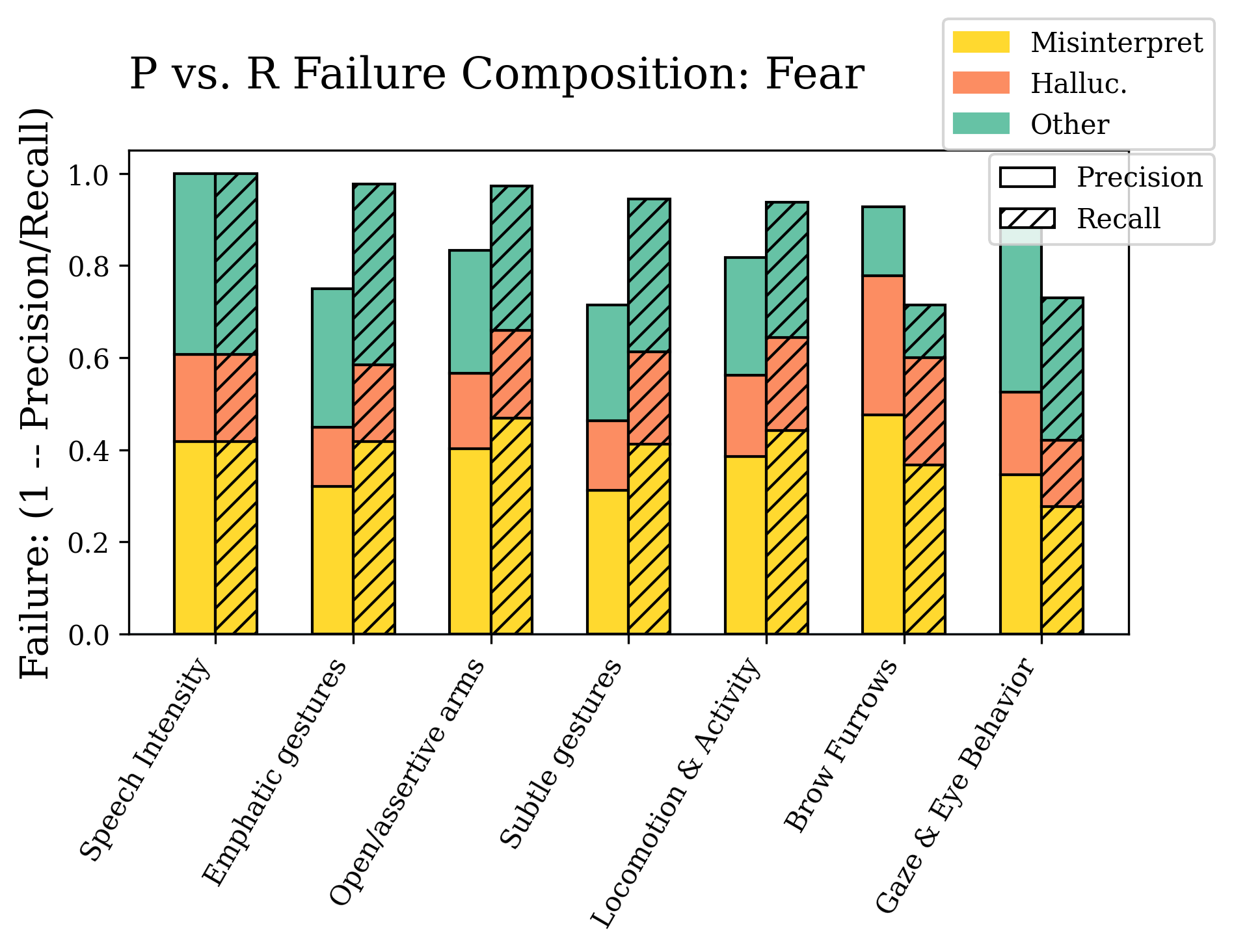}
  \caption{Error analysis for `Fear'. Bigger bars = bigger errors.}
  \label{fig:p-r-fear}
\end{figure}

Figures~\ref{fig:p-r-anger},~\ref{fig:p-r-sadness}, and ~\ref{fig:p-r-fear} show the error breakdowns for the remaining emotions. Emotions of disgust and surprise had too few data points to contain significant results. Among the figures, clusters with too few points to be significant are also excluded.

\section{Prompts}
\label{app:prompts}
Figures~\ref{fig:vlm-speaker}-\ref{fig:llm-ep-tv} contain all prompts used during diagnostic experiments and for dataset creation.

\begin{figure}[h!]
    \centering
    \begin{tcolorbox}[colback=gray!30, colframe=black, boxrule=0.5mm, width=\columnwidth]
    \tiny
        You are given a short video clip from a TV show. There is one main speaker. Throughout the video, think out loud about the mouth movements of the people in the foreground. Based on this, decide who the main speaker is.\\
    \end{tcolorbox}
    \caption{VLM Prompt to get speaker.}
    \label{fig:vlm-speaker}
\end{figure}

\begin{figure}[h!]
    \centering
    \begin{tcolorbox}[colback=gray!30, colframe=black, boxrule=0.5mm, width=\columnwidth]
    \tiny
    In 3-4 sentences, without using any adjectives or emotions, describe the changes in the facial features (eyebrows, forehead, eyes, nose, cheeks, mouth and lips.) of the identified speaker. Do not interpret what they mean.
    \end{tcolorbox}
    \caption{VLM Prompt to get facial expression at Level 1}
    \label{fig:vlm-f-1}
\end{figure}

\begin{figure}[h!]
    \centering
    \begin{tcolorbox}[colback=gray!30, colframe=black, boxrule=0.5mm, width=\columnwidth]
    \tiny
    In 3-4 sentences, describe the facial expressions of the speaker, with an emphasis on the features that hint at the person's emotion. Describe what each of the features indicates about the person's emotional state. What emotion might they be feeling?

    \end{tcolorbox}
    \caption{VLM Prompt to get facial expression at Level 3}
    \label{fig:vlm-f-3}
\end{figure}

\begin{figure}[h!]
    \centering
    \begin{tcolorbox}[colback=gray!30, colframe=black, boxrule=0.5mm, width=\columnwidth]
    \tiny
    We want to decide the emotion of the speaker. The options are joy, sadness, fear, disgust, surprise and anger. In 3-4 sentences, think about the facial expression of the speaker and what they indicate. Based on this, what emotion is the speaker feeling?
    \end{tcolorbox}
    \caption{VLM Prompt to get facial expression at Level 3.}
    \label{fig:vlm-f-5}
\end{figure}

\begin{figure}[h!]
    \centering
    \begin{tcolorbox}[colback=gray!30, colframe=black, boxrule=0.5mm, width=\columnwidth]
    \tiny
    In 3-4 sentences, without using any adjectives or emotions, describe the changes in the body language (arms, hands, legs, torso) of the identified speaker. Do not interpret what they mean.
    \end{tcolorbox}
    \caption{VLM Prompt to get body language at Level 1.}
    \label{fig:vlm-b-1}
\end{figure}

\begin{figure}[h!]
    \centering
    \begin{tcolorbox}[colback=gray!30, colframe=black, boxrule=0.5mm, width=\columnwidth]
    \tiny
    In 3-4 sentences, describe the body language of the speaker, with an emphasis on the features that hint at the person's emotion. Describe what each of the features indicates about the person's emotional state. What emotion might they be feeling?
    \end{tcolorbox}
    \caption{VLM Prompt to get body language at Level 3.}
    \label{fig:vlm-b-3}
\end{figure}

\begin{figure}[h!]
    \centering
    \begin{tcolorbox}[colback=gray!30, colframe=black, boxrule=0.5mm, width=\columnwidth]
    \tiny
    We want to decide the emotion of the speaker. The options are joy, sadness, fear, disgust, surprise and anger. In 3-4 sentences, think about the body language of the speaker and what they indicate. Based on this, what emotion is the speaker feeling?
    \end{tcolorbox}
    \caption{VLM Prompt to get body language at Level 5.}
    \label{fig:vlm-b-5}
\end{figure}

\begin{figure}[h!]
    \centering
    \begin{tcolorbox}[colback=gray!30, colframe=black, boxrule=0.5mm, width=\columnwidth]
    \tiny
    \#\#\# Task Overview:\\
    
    You will be given a piece of text. The text will consist of facts (Eg: she has a smile) and inferences (Eg: indicating she is relaxed). Your task is separate the facts and inferences. Try to pair up the facts and inferences. The same fact can be repeated for multiple inferences (explicitly repeat it). Say "No Fact" if there is no fact with an inference. Follow the guidelines to do this.\\
    
    \#\#\# Strict Guidelines:\\
    
    - Facts are physical traits like smiles, furrowed brows, other physical traits **without adjectives**.\\
       - The fact may come after the inference in some sentences. Example: Her facial expression is one of happiness and contentment, with a smile on her face. Fact: There is a smile on the speaker's face. Inference: The smile suggests happiness and contentment.\\
    \#\# Definition of Inferences:\\
    - Inferences are **what the facts mean or indicate** such as happiness, sadness etc.\\
       - All emotions are inferences.\\
       - ALL adjectives are inferences (in distress, tense, calm, sadly, etc..)\\
       - Any words like "indicates", "suggests", "appears", etc are pointers to inferences.\\
    3 **VERY STRICT RULE**: Do not come up with inferences on your own. Only cluster the information already present in the text.\\

    \#\#\# Examples:\\
    
    \#\# Example text 1:\\
    
    The speaker, the woman driving the car, has a neutral expression with a slight smile, indicating that she is calm and possibly content. Her eyes are focused forward, suggesting that she is engaged in the conversation. The slight smile on her face hints at a positive emotion, such as happiness or satisfaction. Overall, her facial expression suggests that she is in a relaxed and pleasant emotional state.\\
    
    \#\# Example response 1:\\
    
    Information breakdown:\\
    1. The speaker is a woman driving the car.\\
    2. The speaker has a neutral expression with a slight smile, indicating that she is calm and possibly content.\\
    3. The speaker's eyes are focused forward, suggesting that she is engaged in the conversation.\\
    ...\\
    
    5. The speaker's facial expression suggests that she is in a relaxed and pleasant emotional state.\\
    - Fact Part: The speaker's facial expression is clearly visible.\\
    - Inference Part: The expression suggests that she is in a relaxed and pleasant emotional state.\\
    
    \#\# Example text 2:\\
    
    The speaker in the video appears to be feeling joy. Her facial expression is one of 
    ...\\
    3. The speaker seems to be enjoying the conversation and the moment, which indicates a positive and joyful emotion.\\
       - Fact Part: The speaker is participating in a conversation.\\
       - Inference Part: The speaker is enjoying the conversation and the moment, indicating a positive and joyful emotion.\\
    
    \#\#\# Your text:\\
    
    {text}\\
    
    \#\#\# Your response:\\

    \end{tcolorbox}
    \caption{LLM Prompt to break down video descriptions.}
    \label{fig:llm-break-cues}
\end{figure}

\begin{figure}[h!]
    \centering
    \begin{tcolorbox}[colback=gray!30, colframe=black, boxrule=0.5mm, width=\columnwidth]
    \tiny
    \#\#\# Task Overview\\
    You will be given the text of a conversation. Your task is to predict the top emotion of the speaker of the last sentence. The possible emotions are [joy, surprise, anger, fear, disgust, sadness]. Follow the Task Guidelines and the Response Format.\\

    \#\#\# Task Guidelines\\
    - Think out loud about the possible options [joy, surprise, anger, fear, disgust, sadness] using the text. For each emotion, think about whether the last sentence could be an expression of that emotion.\\
    - Finally, output the top emotion according to the Response Format.\\
    
    \#\#\# Response Format\\
    Thinking out loud: My only allowed emotions are [joy, surprise, anger, fear, disgust, sadness]. Based on the text, I think... <your thoughts>. Therefore... <your choice>.\\
    Emotion: <your top emotion>\\

    \#\#\# Conversation\\
    {conversation}\\
    
    \#\#\# Response\\
    \end{tcolorbox}
    \caption{LLM Prompt to classify emotion with Text only.}
    \label{fig:llm-ep-t}
\end{figure}

\begin{figure}[h!]
    \centering
    \begin{tcolorbox}[colback=gray!30, colframe=black, boxrule=0.5mm, width=\columnwidth]
    \tiny
    \#\#\# Task Overview\\
    You will be given the text of a conversation and some visual cues about the main speaker of the last sentence. Your task is to predict the top emotion of the speaker of the last sentence. The possible emotions are [joy, surprise, anger, fear, disgust, sadness]. Follow the Task Guidelines and the Response Format.\\

    \#\#\# Task Guidelines\\
    - Think out loud about the possible options [joy, surprise, anger, fear, disgust, sadness] using the text and then the visual cues. Carefully consider if each emotion could apply.
    - You must pick an emotion.\\
    - Finally, output the top emotion according to the Response Format.\\
    
    \#\#\# Response Format\\
    Thinking out loud: My only allowed emotions are [joy, surprise, anger, fear, disgust, sadness]. Based on the text I think... <your thoughts>. Now, based on the visual cues, I think... <your thoughts>. Therefore... <your choice>.
    Emotion: <your top emotion>\\

    \#\#\# Conversation\\
    {conversation}

    \#\#\# Visual cues\\
    {cues}\\
    
    \#\#\# Response\\
    \end{tcolorbox}
    \caption{LLM Prompt to classify emotion with Text + Vision.}
    \label{fig:llm-ep-tv}
\end{figure}

\begin{figure}[h!]
    \centering
    \begin{tcolorbox}[colback=gray!30, colframe=black, boxrule=0.5mm, width=\columnwidth]
    \tiny
    \#\#\# Task Overview\\
    You will be given some visual cues about a speaker extracted from a video clip. Your task is to predict the top emotion of the speaker. The possible emotions are [joy, surprise, anger, fear, disgust, sadness]. Follow the Task Guidelines and the Response Format.\\

    \#\#\# Task Guidelines\\
    - Think out loud about the possible options [joy, surprise, anger, fear, disgust, sadness] using the visual cues. Carefully consider if each emotion could apply.
    - You must pick an emotion.\\
    - Finally, output the top emotion according to the Response Format.\\
    
    \#\#\# Response Format\\
    Thinking out loud: My only allowed emotions are [joy, surprise, anger, fear, disgust, sadness]. Based on the visual cues, I think... <your thoughts>. Therefore... <your choice>.
    Emotion: <your top emotion>\\

    \#\#\# Visual cues\\
    {cues}\\
    
    \#\#\# Response\\
    \end{tcolorbox}
    \caption{LLM Prompt to classify emotion with Vision.}
    \label{fig:llm-ep-v}
\end{figure}

\section{Implementation Details}
\label{app:implementation}

\begin{table}[h]
\centering
\small
\begin{tabular}{|c|c|}
\hline
\textbf{Emotion 1} & \textbf{Emotion 2} \\
\hline
joy      & anger     \\
joy      & sadness   \\
joy      & fear      \\
joy      & disgust   \\
sadness  & surprise  \\
sadness  & fear      \\
fear     & disgust   \\
sadness  & disgust   \\
anger    & fear      \\
anger    & disgust   \\
\hline
\end{tabular}
\caption{Pairs of emotions that were included in list of "bad" mistakes.}
\label{tab:emotion-bad-combos}
\end{table}

\begin{table}[h]
\centering
\small
\begin{tabular}{|c|c|}
\hline
\textbf{Emotion} & \textbf{Opposite Emotion} \\
\hline
anger     & joy     \\
fear      & joy     \\
joy       & sadness \\
sadness   & joy     \\
disgust   & joy     \\
surprise  & sadness \\
\hline
\end{tabular}
\caption{Emotion-opposite pairs.}
\label{tab:emotion-opposites}
\end{table}

\end{document}